\documentclass{article}

\usepackage{times}
\usepackage{graphicx} 
\usepackage{subfigure} 
\usepackage{natbib}
\usepackage{enumitem}
\usepackage{algorithm}
\usepackage{algorithmic}

\ifdefined\icmlformat
\usepackage{icml2015} 
\else
\usepackage[accepted]{arxiv_icml}
\fi

\usepackage{amsmath,amssymb,amsfonts}
\newtheorem{example}{Example}
\newtheorem{theorem}{Theorem}
\newcommand{\BlackBox}{\rule{1.5ex}{1.5ex}}  
\newenvironment{proof}{\par\noindent{\bf Proof\ }}{\hfill\BlackBox\\[2mm]}

\newcommand{\E}{\mathbb{E}}

\newcommand{\1}{\mathbf{1}}
\def\loss{{\small\texttt{loss}}}
\def\TRAIN{{\small \texttt{TRAIN}}}
\def\TEST{{\small \texttt{TEST}}}
\def\VAL{{\small \texttt{VAL}}}

\def\X{\mathcal{X}}
\def\Y{\mathcal{Y}}
\newcommand{\R}{\mathbb{R}}

\icmltitlerunning{Non-stochastic Best Arm Identification and Hyperparameter Optimization}

\begin{document} 

\twocolumn[
\icmltitle{Non-stochastic Best Arm Identification and Hyperparameter Optimization}

\icmlauthor{Kevin Jamieson}{kgjamieson@wisc.edu}
\icmladdress{Electrical and Computer Engineering Department, University of Wisconsin, Madison, WI 53706}
\icmlauthor{Ameet Talwalkar}{ameet@cs.ucla.edu}
\icmladdress{Computer Science Department, UCLA, Boelter Hall 4732, Los Angeles, CA 90095}

\icmlkeywords{boring formatting information, machine learning, ICML}

\vskip 0.3in
]

\begin{abstract} 
Motivated by the task of hyperparameter optimization,
we introduce the {\em non-stochastic best-arm identification
problem}. Within the multi-armed bandit literature, the cumulative regret objective
enjoys algorithms and analyses for both the non-stochastic and stochastic
settings while to the best of our knowledge, the best-arm identification
framework has only been considered in the stochastic setting. We introduce
the non-stochastic setting under this framework, identify a known algorithm that is
well-suited for this setting, and analyze its behavior. Next, by
leveraging the iterative nature of standard machine learning algorithms, we
cast hyperparameter optimization as an instance of non-stochastic best-arm
identification, and empirically evaluate our proposed algorithm on this task.
Our empirical results show that, by allocating more resources to promising
hyperparameter settings, we typically achieve comparable test accuracies an order of magnitude faster than
baseline methods. 


\end{abstract} 

\section{Introduction}


As supervised learning methods are becoming more widely adopted, hyperparameter
optimization has become increasingly important to simplify and speed up the
development of data processing pipelines while simultaneously yielding more
accurate models.  In hyperparameter optimization for supervised learning, we
are given labeled training data, a set of hyperparameters associated with our
supervised learning methods of interest, and a search space over these
hyperparameters.  We aim to find a particular configuration of hyperparameters
that optimizes some evaluation criterion, e.g., loss on a validation dataset. 

\begin{figure}[h]
\begin{center}
\includegraphics[scale=0.35]{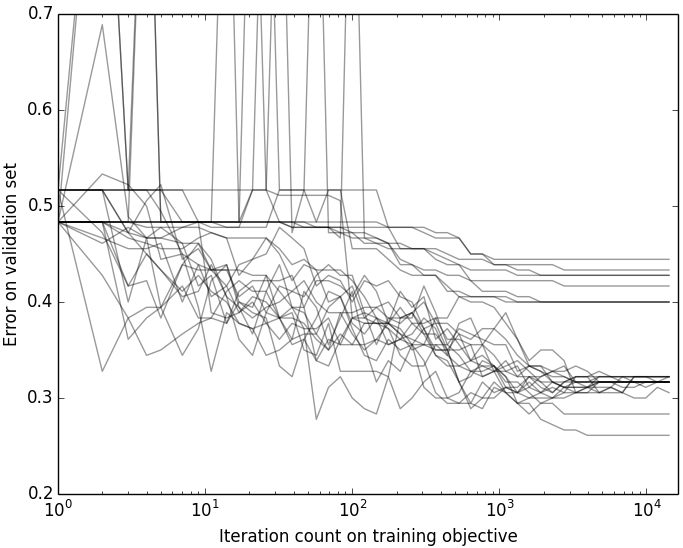}
\caption{Validation error for different hyperparameter choices for a classification task trained
using stochastic gradient descent.}
\label{sourceTraces}
\end{center}
\end{figure}

Since many machine learning algorithms are iterative in nature, particularly when
working at scale, we can evaluate the quality of intermediate results, i.e.,
partially trained learning models, resulting in a sequence of losses that
eventually converges to the final loss value at convergence. For example,
Figure~\ref{sourceTraces} shows the sequence of validation losses for
various hyperparameter settings for kernel SVM models trained via stochastic
gradient descent. The figure shows high variability in model quality across
hyperparameter settings.  It thus seems natural to ask the question: \emph{Can
we terminate these poor-performing hyperparameter settings early in a
principled online fashion to speed up hyperparameter optimization?} 

Although several hyperparameter optimization methods have been proposed
recently, e.g.,~\citet{Snoek2012practical, Snoek2014input,
Hutter2011sequential, Bergstra2011algorithms, Bergstra2012random}, the vast
majority of them consider the training of machine learning models to be
black-box procedures, and only evaluate models after they are fully trained to
convergence.  A few recent works have made attempts to exploit intermediate
results. However, these works either require explicit forms for the convergence
rate behavior of the iterates which is difficult to accurately characterize for
all but the simplest cases \cite{agarwal2012oracle, swersky2014freeze}, or focus
on heuristics lacking theoretical underpinnings~\cite{sparks2015tupaq}.
We build upon these previous works, and in particular study the 
multi-armed bandit formulation proposed in \citet{agarwal2012oracle} and 
\citet{sparks2015tupaq}, where each arm corresponds to a fixed hyperparameter setting,
pulling an arm corresponds to a fixed number of training iterations, and the
loss corresponds to an intermediate loss on some hold-out set. 

We aim to provide a robust, general-purpose, and widely applicable bandit-based
solution to hyperparameter optimization.  Remarkably, however, the existing
multi-armed bandits literature fails to address this natural problem setting: a
\emph{non-stochastic best-arm identification} problem.  While multi-armed bandits is a
thriving area of research, we believe that the existing work fails to
adequately address the two main challenges in this setting:
\vspace{-3mm}
\begin{enumerate}[leftmargin=*]
\setlength\itemsep{0em}
\item We know each arm's sequence of losses eventually converges, but we have
no information about the rate of convergence, and the sequence of losses, like those in
Figure~\ref{sourceTraces}, may exhibit a high degree of non-monotonicity and
non-smoothness.
\item The cost of obtaining the loss of an arm can be disproportionately more
costly than pulling it. For example, in the case of hyperparameter
optimization, computing the validation loss is often drastically more expensive
than performing a single training iteration.
\end{enumerate}
\vspace{-4mm}

We thus study this novel bandit setting, which encompasses the hyperparameter
optimization problem, and analyze an algorithm we identify as being
particularly well-suited for this setting. Moreover, we confirm our theory with
empirical studies that demonstrate an order of magnitude speedups relative to
standard baselines on a number of real-world supervised learning problems and
datasets. 




We note that this bandit setting is quite generally applicable.  While the
problem of hyperparameter optimization inspired this work,
the setting itself encompasses the stochastic best-arm identification
problem \cite{bubeck2009pure}, less-well-behaved stochastic sources like
max-bandits \cite{cicirello2005max}, exhaustive subset selection for feature
extraction, and many optimization problems that ``feel'' like stochastic
best-arm problems but lack the i.i.d. assumptions necessary in that setting.

The remainder of the paper is organized as follows: In
Section~\ref{sec:problem_statement} we present the setting of interest,
 provide a survey of related work, and explain why most existing
algorithms and analyses are not well-suited or applicable for our setting.  We
then study our proposed algorithm in
Section~\ref{sec:proposed_algorithm} in our setting of interest, and analyze
its performance relative to a natural baseline.  We then relate these results
to the problem of hyperparameter optimization in
Section~\ref{hyperParameterTuning}, and present our experimental results in
Section~\ref{sec:results}.

\section{Non-stochastic best arm identification}
\label{sec:problem_statement}

Objective functions for multi-armed bandits problems tend to take on one of two
flavors: 1) best arm identification (or pure exploration) in which one is
interested in identifying the arm with the highest average payoff, and 2)
exploration-versus-exploitation in which we are trying to maximize the
cumulative payoff over time \cite{bubeck2012regret}. While the latter has been analyzed in both the
stochastic and non-stochastic settings, we are unaware of any work that
addresses the best arm objective in the non-stochastic setting, which is our
setting of interest.  Moreover, while related, a strategy that is well-suited
for maximizing cumulative payoff is not necessarily well-suited for the
  best-arm identification task, even in the stochastic
  setting~\cite{bubeck2009pure}.

\begin{figure}[h]
\centerline{
\fbox{\parbox[b]{3.1in}{{\underline{\bf Best Arm Problem for Multi-armed Bandits}}  \\[2pt] \small
{\bf input}: $n$ arms where $\ell_{i,k}$ denotes the loss observed on the $k$th pull of the $i$th arm \\[2pt]
{\bf initialize}: $T_i = 1$ for all $i \in [n]$ \\[2pt]
\textbf{for} $t=1,2,3,\dots$ \\[2pt]
\indent \hspace{.4cm} Algorithm chooses an index $I_t \in [n]$ \\[4pt]
\indent \hspace{.4cm} Loss $\ell_{I_t,T_{I_t}}$ is revealed, $T_{I_t} = T_{I_t} + 1$ \\[4pt]
\indent \hspace{.4cm} Algorithm outputs a recommendation $J_t \in [n]$ \\[4pt] 
\indent \hspace{.4cm} Receive external stopping signal, otherwise continue
}}}
\caption{\label{generalAlgorithm} A generalization of the best arm
problem for multi-armed bandits \cite{bubeck2009pure} that applies to both
the stochastic and non-stochastic settings.}
\end{figure}


The algorithm of Figure~\ref{generalAlgorithm} presents a general form of
the best arm problem for multi-armed bandits.
Intuitively, at each time $t$ the goal is to choose $J_t$ such that the arm
associated with $J_t$ has the lowest loss in some sense. Note that while the
algorithm gets to observe the value for an arbitrary arm $I_t$, the algorithm
is only evaluated on its recommendation $J_t$, that it also chooses
arbitrarily.  This is in contrast to the exploration-versus-exploitation game
where the arm that is played is also the arm that the algorithm is evaluated on,
namely, $I_t$. 


The best-arm identification problems defined below require that the losses be generated
by an oblivious adversary, which essentially means that the loss
sequences are independent of the algorithm's actions. Contrast this with an
adaptive adversary that can adapt future losses based on all the arms that the
algorithm has played up to the current time. If the losses are chosen by an
oblivious adversary then without loss of generality we may assume that all the
losses were generated before the start of the game. 
See \cite{bubeck2012regret} for more info.
We now compare the stochastic and the proposed non-stochastic best-arm identification problems.

\begin{description}
\item \textbf{Stochastic} : For all $i \in [n]$, $k \geq 1$, let
$\ell_{i,k}$ be an i.i.d. sample from a probability distribution
supported on $[0,1]$. For each $i$,  $\E[ \ell_{i,k} ]$ exists
and is equal to some constant $\mu_i$ for all $k \geq 1$. The goal is to
identify $\arg\min_i \mu_i$ while minimizing $\sum_{i=1}^n T_i$. 
\item \textbf{Non-stochastic (proposed in this work)} : For all $i
\in [n]$, $k \geq 1$, let $\ell_{i,k} \in \R$ be generated by an oblivious adversary and assume $\nu_i = \displaystyle\lim_{\tau
\rightarrow \infty} \ell_{i,\tau}$ exists. The goal is to identify $\arg\min_i
\nu_i$ while minimizing $\sum_{i=1}^n T_i$.
\end{description}

These two settings are related in that we can always turn the stochastic
setting into the non-stochastic setting by defining $\ell_{i,T_i} =
\frac{1}{T_i} \sum_{k=1}^{T_i} \ell_{i,T_i}'$ where $ \ell_{i,T_i}'$ are the
losses from the stochastic problem; by the law of large numbers $\lim_{\tau
\rightarrow \infty} \ell_{i,\tau} = \E[\ell_{i,1}']$. In fact, we could do
something similar with other less-well-behaved statistics like the minimum (or maximum) of
the stochastic returns of an arm. As described in \citet{cicirello2005max}, we
can define $\ell_{i,T_i} = \min\{ \ell_{i,1}', \ell_{i,2}',\dots,\ell_{i,T_i}'
\}$, which has a limit since $\ell_{i,t}$ is a bounded, monotonically
decreasing sequence. 


However, the generality of the non-stochastic setting introduces novel
challenges.  In the stochastic setting, if we set $\widehat{\mu}_{i,T_i} =
\frac{1}{T_i} \sum_{k=1}^{T_i} \ell_{i,k}$ then $| \widehat{\mu}_{i,T_i} -
\mu_i | \leq \sqrt{ \frac{\log( 4 n T_i^2) }{2 T_i} }$ for all $i \in [n]$ and
$T_i >0$ by applying Hoeffding's inequality and a union bound.  In contrast,
the non-stochastic setting's assumption that $\lim_{\tau \rightarrow \infty}
\ell_{i,\tau}$ exists implies that there exists a non-increasing function
$\gamma_i$ such that $\left| \ell_{i,t} - \lim_{\tau \rightarrow \infty}
\ell_{i,\tau}  \right|  \leq \gamma_i(t)$ and that $\lim_{t \rightarrow \infty}
\gamma_i(t) = 0$. However, the existence of this limit tells us nothing about
how {\em quickly} $\gamma_i(t)$ approaches $0$.  The lack of an explicit
convergence rate as a function of $t$ presents a problem as even the tightest
$\gamma_i(t)$ could decay arbitrarily slowly and we would never know it. 

This observation has two consequences.  First, we can never {\em reject} the
possibility that an arm is the ``best'' arm.  Second, we can never {\em verify}
that an arm is the ``best'' arm or even attain a value within $\epsilon$ of the
best arm.  Despite these challenges, in Section~\ref{sec:proposed_algorithm} we
identify an effective algorithm under natural measures of performance, using
ideas inspired by the fixed budget setting of the stochastic best arm problem
\cite{karnin2013almost,audibert2010best,gabillon2012best}.

\subsection{Related work}
\label{ssec:related}
%

Despite dating to back to the late 1950's, the best-arm identification problem
for the stochastic setting has experienced a surge of activity in the last
decade. The work has two major branches: the fixed budget setting and the
fixed confidence setting. In the fixed budget setting, the algorithm is given
a set of arms and a budget $B$ and is tasked with maximizing the probability
of identifying the best arm by pulling arms without exceeding the total
budget. While these algorithms were developed for and analyzed in the
stochastic setting, they exhibit attributes that are very amenable to the
non-stochastic setting. In fact, the algorithm we propose to use in this
paper is exactly the Successive Halving algorithm of \citet{karnin2013almost},
though the non-stochastic setting requires its own novel analysis that we present
in Section~\ref{sec:proposed_algorithm}.
Successive Rejects \cite{audibert2010best} is another fixed budget algorithm
that we compare to in our experiments.    

The best-arm identification problem in the fixed confidence setting takes an
input $\delta \in (0,1)$ and guarantees to output the best arm with probability
at least $1-\delta$ while attempting to minimize the number of total arm pulls.
These algorithms rely on probability theory to determine how many times each
arm must be pulled in order to decide if the arm is suboptimal and
should no longer be pulled, either by explicitly discarding it, e.g.,
Successive Elimination~\cite{even2006action} and Exponential Gap Elimination
\cite{karnin2013almost}, or implicitly by other methods, e.g., LUCB
\cite{kalyanakrishnan2012pac} and Lil'UCB \cite{jamieson2014lil}. Algorithms from
the fixed confidence setting are ill-suited for the non-stochastic best-arm
identification problem because they rely on statistical bounds that are generally
not applicable in the non-stochastic case. These algorithms also exhibit some
undesirable behavior with respect to how many losses they observe, which we
explore next.

\begin{table}
\centering
\begin{tabular}{ l | c }
Exploration algorithm & $\#$ observed losses \\
\hline
Uniform (baseline) (B) & $n$ \\
Successive Halving* (B)  & $2n+1$ \\
Successive Rejects (B) & $(n+1)n/2$ \\
Successive Elimination (C) & $n \log_2(2B)$ \\
LUCB (C), lil'UCB (C), EXP3 (R) & $B$ \\
\end{tabular}
\caption{The number of times an algorithm observes a loss in
terms of budget $B$ and number of arms $n$, where $B$ is known to
the algorithm. (B), (C), or (R) indicate whether the algorithm is of the fixed
budget, fixed confidence, or cumulative regret variety, respectfully. (*) indicates
the algorithm we propose for use in the non-stochastic best arm setting.}
\label{fevalsTable}
\end{table}

In addition to just the total number of arm pulls, this work also considers the
required number of {\em observed} losses. This is a natural cost to consider
when $\ell_{i,T_i}$ for any $i$ is the result of doing some computation like
evaluating a partially trained classifier on a hold-out validation set or
releasing a product to the market to probe for demand. In some cases the cost, be it time, effort, or dollars, of an evaluation of the loss of an arm
after some number of pulls can dwarf the cost of pulling the arm. Assuming a known time horizon (or budget), Table~\ref{fevalsTable} describes the total number of
times various algorithms observe a loss as a function of the budget $B$ and the
number of arms $n$. We include in our comparison the EXP3 algorithm \cite{auer2002nonstochastic}, a
popular approach for minimizing cumulative regret in the non-stochastic setting.  In practice $B \gg n$,
and thus Successive Halving is a particular attractive option, as along with
the baseline, it is the only algorithm that observes losses proportional to the
number of arms and independent of the budget. As we will see in
Section~\ref{sec:results}, the performance of these algorithms is quite
dependent on the number of observed losses.

\section{Proposed algorithm and analysis}
\label{sec:proposed_algorithm}

The proposed Successive Halving algorithm of Figure~\ref{succHalfAlg} was
originally proposed for the stochastic best arm identification problem in the
fixed budget setting by \cite{karnin2013almost}. However, our novel analysis in
this work shows that it is also effective in the non-stochastic setting.  The
idea behind the algorithm is simple: given an input budget, uniformly allocate
the budget to a set of arms for a predefined amount of iterations, evaluate
their performance, throw out the worst half, and repeat until just one arm
remains. 

 \begin{figure}[h]
\centerline{
\fbox{\parbox[b]{3.1in}{{\underline{\bf Successive Halving Algorithm}}  \\[2pt] \small
{\bf input}: Budget $B$, $n$ arms where $\ell_{i,k}$ denotes the $k$th loss from the $i$th arm \\[2pt]
{\bf Initialize}: $S_0 = [n]$. \\[2pt] 
\textbf{For} $k=0,1,\dots,\lceil \log_2(n) \rceil -1$ \\
\indent \hspace{.4cm} Pull each arm in $S_k$ for $r_k = \lfloor \frac{ B }{|S_k| \lceil \log_2(n) \rceil } \rfloor$ additional times and set $R_k = \sum_{j=0}^k r_j$.\\[4pt]
\indent \hspace{.4cm} Let $\sigma_k$ be a bijection on $S_k$ such that $\ell_{\sigma_k(1) , R_k}  \leq \ell_{\sigma_k(2) , R_k}  \leq \dots \leq \ell_{\sigma_k(|S_k|) , R_k}  $ \\[2pt]      
\indent \hspace{.4cm} $\displaystyle S_{k+1} = \left\{ i \in S_k : \ell_{\sigma_k(i) , R_k} \leq \ell_{\sigma_k(  \lfloor |S_k| /2 \rfloor  ) , R_k}   \right\}$.\\[2pt]
{\bf output} : Singleton element of $S_{\lceil \log_2(n) \rceil}$
}}
}
\caption{Successive Halving was originally proposed for the
stochastic best arm identification problem in 
\citet{karnin2013almost} but is also applicable to the non-stochastic setting.}
\label{succHalfAlg}
\end{figure}
The budget as an input is easily removed by the ``doubling trick'' that
attempts $B \leftarrow n$, then $B \leftarrow 2 B$, and so on. This method can
reuse existing progress from iteration to iteration and effectively makes the
algorithm parameter free. But its most notable quality is that if a budget of
$B'$ is necessary to succeed in finding the best arm, by performing the
doubling trick one will have only had to use a budget of $2B'$ in the worst
case without ever having to know $B'$ in the first place. Thus, for the
remainder of this section we consider a fixed budget.

\subsection{Analysis of Successive Halving}

We first show that the algorithm never takes a total number of samples
that exceeds the budget $B$:
\begin{align*}
\sum_{k=0}^{\lceil \log_2(n) \rceil -1} |S_k|  \left\lfloor \tfrac{ B }{|S_k| \lceil \log(n) \rceil } \right\rfloor \leq \sum_{k=0}^{\lceil \log_2(n) \rceil -1} \tfrac{ B }{\lceil \log(n) \rceil }  \leq B \, .
\end{align*}
Next we consider how the algorithm performs in terms of identifying the best
arm. First, for $i=1,\dots,n$ define $\nu_i = \lim_{\tau \rightarrow \infty}
\ell_{i,\tau}$ which exists by assumption. Without loss of generality, assume
that
\begin{align*}
\nu_1 < \nu_2 \leq \dots \leq \nu_n \,.
\end{align*}
We next introduce functions that bound the approximation error of $\ell_{i,t}$
with respect to $\nu_i$ as a function of $t$. For each $i =1,2,\dots,n$ let
$\gamma_i(t)$ be the point-wise smallest, non-increasing function of $t$ such
that 
\begin{align*}
| \ell_{i,t} - \nu_i | \leq \gamma_{i}(t) \quad \forall t.
\end{align*} 
In addition, define ${\gamma}_i^{-1}( \alpha ) = \min \{ t \in \mathbb{N} :
\gamma_i(t) \leq \alpha \}$ for all $i \in [n]$. With this definition, if $t_i > {\gamma}_i^{-1}( \frac{\nu_i -
\nu_1}{2} )$ and $t_1 >  {\gamma}_1^{-1}( \frac{\nu_i - \nu_1}{2} )$ then
\begin{align*}
\ell_{i,t_i} - \ell_{1,t_1} &= (\ell_{i,t_i} - \nu_i) + (\nu_1 - \ell_{1,t_1}) + 2 \big (\tfrac{\nu_i - \nu_1}{2}\big )\\
&\geq -\gamma_i(t_i) - \gamma_1(t_1) +  2 \big (\tfrac{\nu_i - \nu_1}{2}\big )  > 0.
\end{align*}
Indeed, if $\min\{t_i,t_1\} > \max\{ {\gamma}_i^{-1}( \frac{\nu_i
-\nu_1}{2} ), {\gamma}_1^{-1}( \frac{\nu_i - \nu_1}{2} ) \}$ then we are
guaranteed to have that $\ell_{i,t_i} > \ell_{1,t_1}$. That is, comparing the
intermediate values at $t_i$ and $t_1$ suffices to determine the ordering of the
final values $\nu_i$ and $\nu_1$. Intuitively, this condition holds because 
the envelopes at the given times, namely
$\gamma_i(t_i)$ and $\gamma_1(t_1)$, are small relative to the gap between $\nu_i$ and $\nu_1$. This line of reasoning is at the heart of the
proof of our main result, and the theorem is stated in terms of these
quantities. All proofs can be found in the appendix.
\begin{theorem} \label{adaptive}
Let $\nu_i = \displaystyle\lim_{\tau \rightarrow \infty} \ell_{i,\tau}$, $\displaystyle\bar{\gamma}(t) = \max_{i=1,\dots,n} \gamma_i(t)$ and 
\begin{align*}
z &=  2 \lceil \log_2(n) \rceil   \, \max_{i=2,\dots,n}  i \, ( 1 + \bar{\gamma}^{-1} \left(\tfrac{ \nu_{i} - \nu_1 }{2} \right) )  \\
&\leq 2 \lceil \log_2(n) \rceil  \big( n + \sum_{i=2,\dots,n}  \bar{\gamma}^{-1} \left( \tfrac{  \nu_i  -\nu_1 }{2} \right) \big).
\end{align*} 
If the budget $B>z$ then the best arm is returned from the algorithm. 
\end{theorem}

The representation of $z$ on the right-hand-side of the inequality is very
intuitive: if $\bar{\gamma}(t) = \gamma_i(t) \ \ \forall i$ and an oracle gave us
an explicit form for $\bar{\gamma}(t)$, then to merely verify that the $i$th arm's
final value is higher than the best arm's, one must pull each of the two arms
at least a number of times equal to the $i$th term in the sum (this becomes
clear by inspecting the proof of Theorem~\ref{nonadaptive_necessity}).
Repeating this argument for all $i=2,\dots,n$ explains the sum over all $n-1$
arms. While clearly not a proof, this argument along with known lower bounds
for the stochastic setting \cite{audibert2010best,kaufmann2014complexity}, a
subset of the non-stochastic setting, suggest that the above result may be
nearly tight in a minimax sense up to $\log$ factors.


\begin{example}
Consider a feature-selection problem where you are given a dataset $\{
(x_i,y_i) \}_{i=1}^n$ where each $x_i \in \R^D$ and you are tasked with
identifying the best subset of features of size $d$ that linearly predicts
$y_i$ in terms of the least-squares metric. In our framework, each $d$-subset
is an arm and there are $n = \binom{D}{d}$ arms. Least squares is a convex
quadratic optimization problem that can be efficiently solved with stochastic
gradient descent. Using known bounds for the rates of convergence
\cite{nemirovski2009robust} one can show that $\gamma_a(t) \leq \frac{\sigma_a
\log(nt/\delta)}{t}$ for all $a=1,\dots,n$ arms and all $t\geq1$ with probability at least $1-\delta$ where $\sigma_a$ is a
constant that depends on the condition number of the quadratic defined by the
$d$-subset. Then in Theorem~\ref{adaptive}, $\bar{\gamma}(t) = \frac{\sigma_{\max}
\log(nt/\delta)}{t}$ with  $\sigma_{\max} = \max_{a=1,\dots,n} \sigma_a$ so after inverting $\bar{\gamma}$ we find that $z = 2 \lceil \log_2(n) \rceil \max_{a=2,\dots,n} a  \frac{ 4 \sigma_{\max} \log\left( \tfrac{ 2n \sigma_{\max} }{ \delta (\nu_a - \nu_1)}  \right) }{\nu_a - \nu_1}$ is a sufficient budget to identify the best arm. Later we put this result in context by comparing to a baseline strategy. 
\end{example}

In the above example we computed upper bounds on the $\gamma_i$ functions in terms of problem dependent parameters to provide us with a sample complexity by plugging these values into our theorem. However, we stress that constructing tight bounds for the $\gamma_i$ functions is very difficult outside of very simple problems like the one described above, and even then we have unspecified constants. Fortunately, because our algorithm is agnostic to these $\gamma_i$ functions, it is also in some sense {\em adaptive} to them: the faster the arms' losses converge, the faster the best arm is discovered, without ever changing the algorithm. This behavior is in stark contrast to the hyperparameter tuning work of \citet{swersky2014freeze} and \citet{agarwal2012oracle}, in which the algorithms explicitly take upper bounds on these $\gamma_i$ functions as input, meaning the performance of the algorithm is only as good as the tightness of these difficult to calculate bounds.

\subsection{Comparison to a uniform allocation strategy}
We can also derive a result for the naive uniform budget allocation strategy.
For simplicity, let $B$ be a multiple of $n$ so that at the end
  of the budget we have $T_i = B/n$ for all $i \in [n]$ and the output arm is
  equal to $\widehat{i} = \arg\min_{i} \ell_{i,B/n}$. 

\begin{theorem} \label{nonadaptive}
(Uniform strategy -- sufficiency) Let $\nu_i = \displaystyle\lim_{\tau \rightarrow \infty} \ell_{i,\tau}$, $\bar{\gamma}(t) = \max_{i=1,\dots,n} \gamma_i(t)$ and 
\begin{align*}
z &=\max_{i=2,\dots,n} n  \bar{\gamma}^{-1} \left( \tfrac{ \nu_{i} - \nu_1}{2} \right).
\end{align*} 
If $B>z$ then the uniform strategy returns the best arm. 
\end{theorem}

Theorem~\ref{nonadaptive} is just a sufficiency statement so it is unclear how the performance of the method actually compares to the Successive Halving result of Theorem~\ref{adaptive}. The next theorem says that the above result is tight in a worst-case sense, exposing the real gap between the algorithm of Figure~\ref{succHalfAlg} and the naive uniform allocation strategy. 

\begin{theorem} \label{nonadaptive_necessity}
(Uniform strategy -- necessity)
For any given budget $B$ and final values $\nu_1 < \nu_2 \leq \dots \leq \nu_n$ there exists a sequence of losses $\{ \ell_{i,t} \}_{t=1}^\infty$, $i=1,2,\dots,n$ such that if
\begin{align*}  
B < \max_{i=2,\dots,n} n  \bar{\gamma}^{-1} \left( \tfrac{ \nu_{i} - \nu_1}{2} \right)
\end{align*}
then the uniform budget allocation strategy will not return the best arm. 
\end{theorem}

If we consider the second, looser representation of $z$ on the right-hand-side of the inequality in Theorem~\ref{adaptive} and multiply this quantity by $\frac{n-1}{n-1}$ we see that the sufficient number of pulls for the Successive Halving algorithm essentially behaves like $(n-1) \log_2(n)$ times the {\em average}   $ \frac{1}{n-1} \sum_{i=2,\dots,n}  \bar{\gamma}^{-1} \left( \frac{  \nu_i  -\nu_1 }{2} \right)$ whereas the necessary result of the uniform allocation strategy of Theorem~\ref{nonadaptive_necessity} behaves like $n$ times the {\em maximum}   $ \max_{i=2,\dots,n}  \bar{\gamma}^{-1} \left( \frac{ \nu_{i} - \nu_1}{2} \right)$. The next example shows that the difference between this average and max can be very significant.

\begin{example}
Recall Example 1 and now assume that $\sigma_a = \sigma_{\max}$ for all
$a=1,\dots,n$. Then Theorem~\ref{nonadaptive_necessity} says that the uniform
allocation budget must be at least $n  \frac{ 4 \sigma_{\max} \log\left(
\tfrac{ 2n \sigma_{\max} }{ \delta (\nu_2 - \nu_1)}  \right) }{\nu_2 - \nu_1}$
to identify the best arm. To see how this result compares with
that of 
Successive Halving, let us
parameterize the $\nu_a$ limiting values such that $\nu_a = a/n$ for
$a=1,\dots,n$. Then a sufficient budget for the Successive Halving algorithm to
identify the best arm is just
$8 n \lceil \log_2(n) \rceil  \sigma_{\max} \log\left( \tfrac{ n^2 \sigma_{\max} }{ \delta}  \right)$ 
while the uniform allocation strategy would require a budget of at least
$2 n^2   \sigma_{\max} \log\left( \tfrac{ n^2 \sigma_{\max} }{ \delta}  \right)$. This is 
a difference of essentially $4 n\log_2(n)$ versus $n^2$.
\end{example}

\subsection{A pretty good arm} \label{prettyGoodArm}

Up to this point we have been concerned with identifying the {\em best} arm:
$\nu_1 = \arg\min_i \nu_i$ where we recall that $\nu_i =
\displaystyle\lim_{\tau \rightarrow \infty} \ell_{i,\tau}$. But in practice one
may be satisfied with merely an $\epsilon$-good arm $i_\epsilon$ in the sense
that $\nu_{i_\epsilon} - \nu_1 \leq \epsilon$. However, with our minimal
assumptions, such a statement is impossible to make since we have no knowledge
of the $\gamma_i$ functions to determine that an arm's final value is within
$\epsilon$ of any value, much less the unknown final converged value of the
best arm. 
However, as we show in Theorem~\ref{thm:prettygood}, the Successive
Halving algorithm cannot do much worse than the uniform allocation strategy.  

\begin{theorem} \label{thm:prettygood}
For a budget $B$ and set of $n$ arms, define $\widehat{i}_{SH}$ as the
output of the Successive Halving algorithm. Then
\begin{align*}   
\nu_{\widehat{i}_{SH}} &- \nu_1 \leq \lceil\log_2(n)\rceil 2 \bar{\gamma}\left(  \lfloor \tfrac{ B }{n \lceil \log_2(n) \rceil } \rfloor \right).
\end{align*}
Moreover, $\widehat{i}_{U}$, the
output of the uniform strategy, satisfies 
\begin{align*}
\nu_{\widehat{i}_U} - \nu_1 \leq  \ell_{\widehat{i},B/n} - \ell_{1,B/n}  + 2 \bar{\gamma}(B/n) \leq 2 \bar{\gamma}(B/n).
\end{align*}
\end{theorem}

\begin{example}
Recall Example~1. Both the Successive Halving algorithm and the uniform allocation strategy satisfy 
$\widehat{\nu}_i - \nu_1 \leq \widetilde{O}\left( n/B\right)$ where $\widehat{i}$ is the output of either algorithm and $\widetilde{O}$ suppresses $\text{poly}\log$ factors. 
\end{example}
We stress that this result is merely a fall-back guarantee, ensuring that we
can never do much worse than uniform. However, it does not rule out the
possibility of the Successive Halving algorithm far outperforming the uniform
allocation strategy in practice. Indeed, we observe order of magnitude speed
ups in our experimental results. 

\section{Hyperparameter optimization for supervised learning} \label{hyperParameterTuning}

In supervised learning we are given a dataset that is composed of pairs
$(x_i,y_i) \in \X \times \Y$ for $i=1,\dots,n$ sampled i.i.d. from some unknown
joint distribution $P_{X,Y}$, and we are tasked with finding a map (or model)
$f : \X \rightarrow \Y$ that minimizes $\E_{(X,Y) \sim P_{X,Y}}\left[ \loss(
f(X),Y ) \right]$ for some known loss function $\loss : \Y \times \Y
\rightarrow \R$. Since $P_{X,Y}$ is unknown, we cannot compute $\E_{(X,Y)\sim
P_{XY}}[ \loss( f(X),Y ) ]$ directly, but given $m$ additional samples drawn
i.i.d. from $P_{X,Y}$ we can approximate it with an empirical estimate, that
is, $\frac{1}{m} \sum_{i=1}^{m} \loss( f(x_i),y_i)$. We do not consider
arbitrary mappings $\X \rightarrow \Y$ but only those that are the output of
running a fixed, possibly randomized, algorithm $\mathcal{A}$ that takes a
dataset $\{ (x_i,y_i) \}_{i=1}^n$ and algorithm-specific parameters $\theta \in
\Theta$ as input so that for any $\theta$ we have $f_\theta =\mathcal{A}\left(
\{ (x_i,y_i) \}_{i=1}^n,  \theta \right)$ where $f_\theta : \X \rightarrow \Y$.
For a fixed dataset $\{ (x_i,y_i) \}_{i=1}^n$ the parameters $\theta \in
\Theta$ index the different functions $f_\theta$, and will henceforth be
referred to as {\em hyperparameters}. We adopt the train-validate-test
framework for choosing hyperparameters \cite{hastie2005elements}: 
\begin{enumerate}
\setlength\itemsep{-.25em}
\item Partition the total dataset into {\small \texttt{TRAIN}, \texttt{VAL} }, and \texttt{TEST} sets with $\texttt{TRAIN} \cup \texttt{VAL} \cup \texttt{TEST} = \{ (x_i,y_i) \}_{i=1}^m$. 
\item Use \texttt{TRAIN} to train a model  $f_\theta =\mathcal{A}\left( \{ (x_i,y_i) \}_{i\in \texttt{TRAIN}},  \theta \right)$ for each $\theta \in \Theta$,
\item Choose the hyperparameters that minimize the empirical loss on the examples in
\texttt{VAL}: $\widehat{\theta} = \arg\min_{\theta \in \Theta}
\tfrac{1}{|\texttt{VAL} |} \sum_{i \in \texttt{VAL} } \loss(
f_\theta(x_i),y_i)$
\item Report the empirical loss of $\widehat{\theta}$ on the test error: $ \tfrac{1}{|\texttt{TEST} |} \sum_{i \in \texttt{TEST} } \loss( f_{\widehat{\theta}} (x_i),y_i) $.\end{enumerate}
 \begin{example}
Consider a linear classification example where $\X \times \Y = \R^d \times \{-1,1\}$, $\Theta \subset \R_+$,  $f_\theta = \mathcal{A}\left( \{ (x_i,y_i) \}_{i \in \texttt{TRAIN}}, \theta \right)$ where $f_\theta(x) = \langle w_\theta , x \rangle$ with $w_\theta= \arg\min_{w} \frac{1}{|\texttt{TRAIN}|} \sum_{i \in \texttt{TRAIN}} \max( 0 , 1- y_i \langle w, x_i \rangle) + \theta ||w||_2^2$, and finally $\widehat{\theta} = \arg\min_{\theta \in \Theta}  \frac{1}{ |\texttt{VAL}| } \sum_{i \in \texttt{VAL}}  \1\{  y \, f_\theta(x) < 0 \} $. 
\end{example}
In the simple above example involving a single hyperparameter, we emphasize
that for each $\theta$ we have that $f_\theta$ can be efficiently computed
using an iterative algorithm \cite{shalev2011pegasos}, however, the selection
of $\widehat{f}$ is the minimization of a function that is not necessarily even
continuous, much less convex. This pattern is more often the rule than the
exception.
We next attempt to generalize and exploit this observation.
\subsection{Posing as a best arm non-stochastic bandits problem}
\label{parameterTuningAsBandits}
Let us assume that the algorithm $\mathcal{A}$ is iterative so that
for a given $ \{ (x_i,y_i) \}_{i \in \texttt{TRAIN}}$ and $\theta$, 
the algorithm outputs a function $f_{\theta,t}$ every iteration $t >  1$ and
we may compute
\begin{align*}
\ell_{\theta,t} =  \tfrac{1}{|\texttt{VAL} |} \sum_{i \in \texttt{VAL} } \loss( f_{\theta,t} (x_i),y_i) .
\end{align*}
We assume that the limit $\lim_{t \rightarrow \infty} \ell_{\theta,t}$
exists\footnote{We note that $f_\theta = \lim_{t \rightarrow \infty}
f_{\theta,t}$ is not enough to conclude that $\lim_{t \rightarrow \infty}
\ell_{\theta,t}$ exists (for instance, for classification with $0/1$ loss this
is not necessarily true) but these technical issues can usually be usurped for
real datasets and losses (for instance, by replacing $\1\{ z <0\}$ with
a very steep sigmoid). We ignore this technicality in our experiments.} and is equal to $
\frac{1}{|\texttt{VAL} |} \sum_{i \in \texttt{VAL} } \loss( f_{\theta}
(x_i),y_i) $.  

With this transformation we are in the position to put the hyperparameter
optimization problem into the framework of Figure~\ref{generalAlgorithm} and,
namely, the non-stochastic best-arm identification formulation developed in the above
sections. We generate the arms (different hyperparameter settings) uniformly
at random (possibly on a log scale) from within the region of valid
hyperparameters (i.e. all hyperparameters within some minimum and maximum
ranges) and sample enough arms to ensure a sufficient cover of the
space~\cite{Bergstra2012random}.  Alternatively, one could input a uniform grid
over the parameters of interest.  We note that random search and grid search
remain the default choices for many open source machine learning packages such
as LibSVM~\cite{Chang01libsvm}, scikit-learn~\cite{scikit-learn} and
MLlib~\cite{Kraska13}. 
As described in Figure~\ref{generalAlgorithm}, the bandit algorithm 
will choose $I_t$, and we will use the
convention that $J_t = \arg\min_{\theta} \ell_{\theta,T_\theta}$. The arm
selected by $J_t$ will be evaluated on the test set following the work-flow
introduced above. 

\subsection{Related work}






We aim to leverage the iterative nature of standard
machine learning algorithms to speed up hyperparameter optimization in a robust
and principled fashion. We now review related work 
in the context of our results.
In Section~\ref{prettyGoodArm} we show that no algorithm
can provably identify a hyperparameter with a value within $\epsilon$ of the
optimal without known, explicit functions $\gamma_i$, which means no algorithm
can reject a hyperparameter setting with absolute confidence without making
potentially unrealistic assumptions. 
\citet{swersky2014freeze} explicitly defines the $\gamma_i$ functions in an
ad-hoc, algorithm-specific, and data-specific fashion which leads to strong
$\epsilon$-good claims.  
A related line of work explicitly defines $\gamma_i$-like functions for
optimizing the computational efficiency of structural risk minimization,
yielding bounds \cite{agarwal2012oracle}. We stress that these
results are only as good as the tightness and correctness of the $\gamma_i$
bounds, and we view our work as an empirical, data-driven driven approach
to the pursuits of \citet{agarwal2012oracle}. Also, 
\citet{sparks2015tupaq} empirically studies an early stopping heuristic for
hyperparameter optimization similar in spirit to the Successive
Halving algorithm.  

We further note that we fix the hyperparameter settings (or
arms) under consideration and adaptively allocate our budget to each arm.
In contrast, Bayesian
optimization advocates choosing hyperparameter settings adaptively, but with
the exception of~\citet{swersky2014freeze}, allocates a
fixed budget to each selected hyperparameter setting~\cite{Snoek2012practical,
Snoek2014input, Hutter2011sequential, Bergstra2011algorithms,
Bergstra2012random}. These Bayesian optimization methods, though heuristic in
nature as they attempt to simultaneously fit and optimize a non-convex and
potentially high-dimensional function, yield promising empirical results.
We view our approach as complementary and orthogonal to the method
used for choosing hyperparameter settings, and extending our approach in a
principled fashion to adaptively choose arms, e.g., in a mini-batch setting, is
an interesting avenue for future work.

\section{Experiment results}
\label{sec:results}
\begin{figure*}[!ht]
\centering
\begin{tabular}{c c}
\includegraphics[width=.4\textwidth]{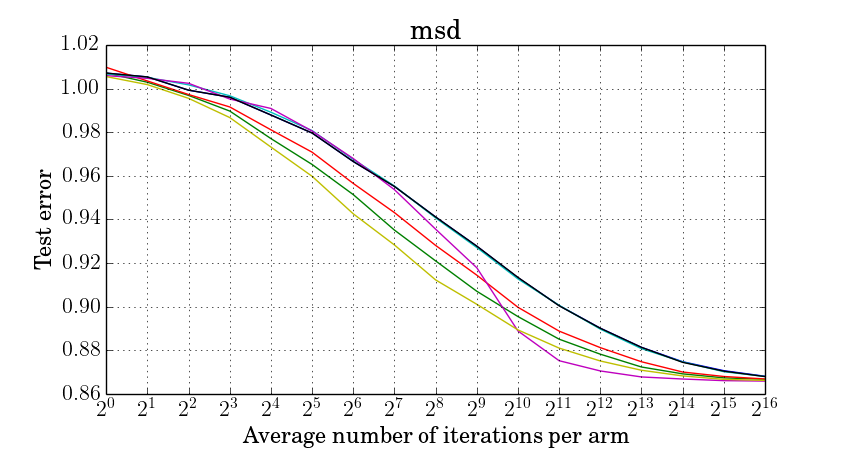} \qquad & \qquad
\includegraphics[width=.4\textwidth]{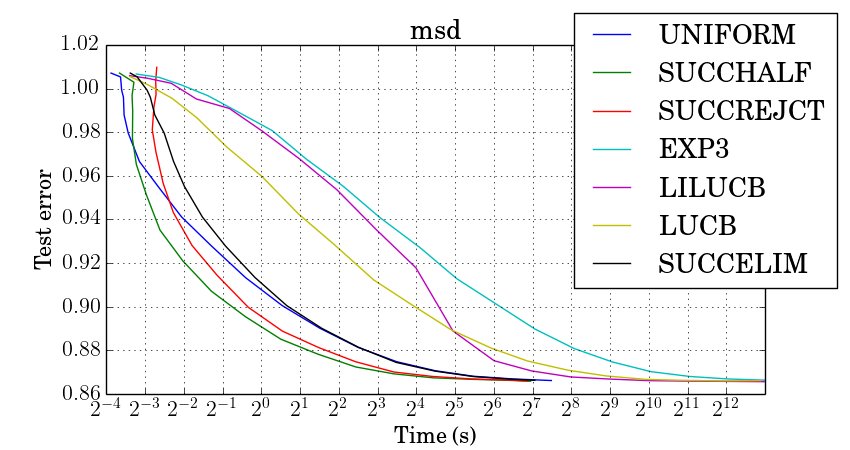}
\end{tabular}
\caption{Ridge Regression. Test error with respect to both the number of iterations (left) and wall-clock time (right). Note that in the left plot, uniform, EXP3, and Successive Elimination are plotted on top of each other.}
 \label{fig:rdgreg}
 \vspace{-.1in}
\end{figure*}


In this section we compare the proposed algorithm to a number of other
algorithms, including the baseline uniform allocation strategy, on a number of
supervised learning hyperparameter optimization problems using the experimental
setup outlined in Section~\ref{parameterTuningAsBandits}. Each experiment was
implemented in Python and run in parallel using the multiprocessing library on
an Amazon EC2 c3.8xlarge instance with 32 cores and 60 GB of memory. In all
cases, full datasets were partitioned into a training-base dataset and a test
(\TEST) dataset with a 90/10 split. The training-base dataset was then
partitioned into a training (\TRAIN) and validation (\VAL) datasets with an
80/20 split. All plots report loss on the test error. 

To evaluate the different search algorithms' performance, we fix a total budget
of iterations and allow the search algorithms to decide how to divide it up
amongst the different arms. The curves are produced by implementing the
doubling trick by simply doubling the measurement budget each time. For the
purpose of interpretability, we reset all iteration counters to 0 at each
doubling of the budget, i.e., we do not warm start upon doubling. All datasets, aside from
the collaborative filtering experiments, are normalized so that each dimension
has mean 0 and variance 1.

\subsection{Ridge regression}

We first consider a ridge regression problem trained with stochastic gradient
descent on this objective function with step size $.01/\sqrt{2 + T_\lambda}$.
The $\ell_2$ penalty hyperparameter $\lambda \in [10^{-6},10^0]$ was chosen
uniformly at random on a log scale per trial, wth $10$ values (i.e., arms) selected per trial. We use
the Million Song Dataset year prediction task \cite{Lichman2013} where we have
down sampled the dataset by a factor of 10 and normalized the years such that
they are mean zero and variance 1 with respect to the training set. The
experiment was repeated for 32 trials. Error on the $\VAL$ and $\TEST$ was
calculated using mean-squared-error.  In the left panel of
Figure~\ref{fig:rdgreg} we note that LUCB, lil'UCB perform the best in the
sense that they achieve a small test error two to four times faster, in terms
of iterations, than most other methods. However, in the right panel the same
data is plotted but with respect to wall-clock time rather than iterations and
we now observe that Successive Halving and Successive Rejects are the top
performers. This is explainable by Table~\ref{fevalsTable}: EXP3, lil'UCB, and
LUCB must evaluate the validation loss on every iteration requiring much
greater compute time. This pattern is observed in all experiments so in the sequel
we only consider the uniform allocation, Successive Halving, and
Successive Rejects algorithm.


\subsection{Kernel SVM}
\begin{figure}[!ht]
\vspace{-.1in}
\centering
\includegraphics[width=.475\textwidth]{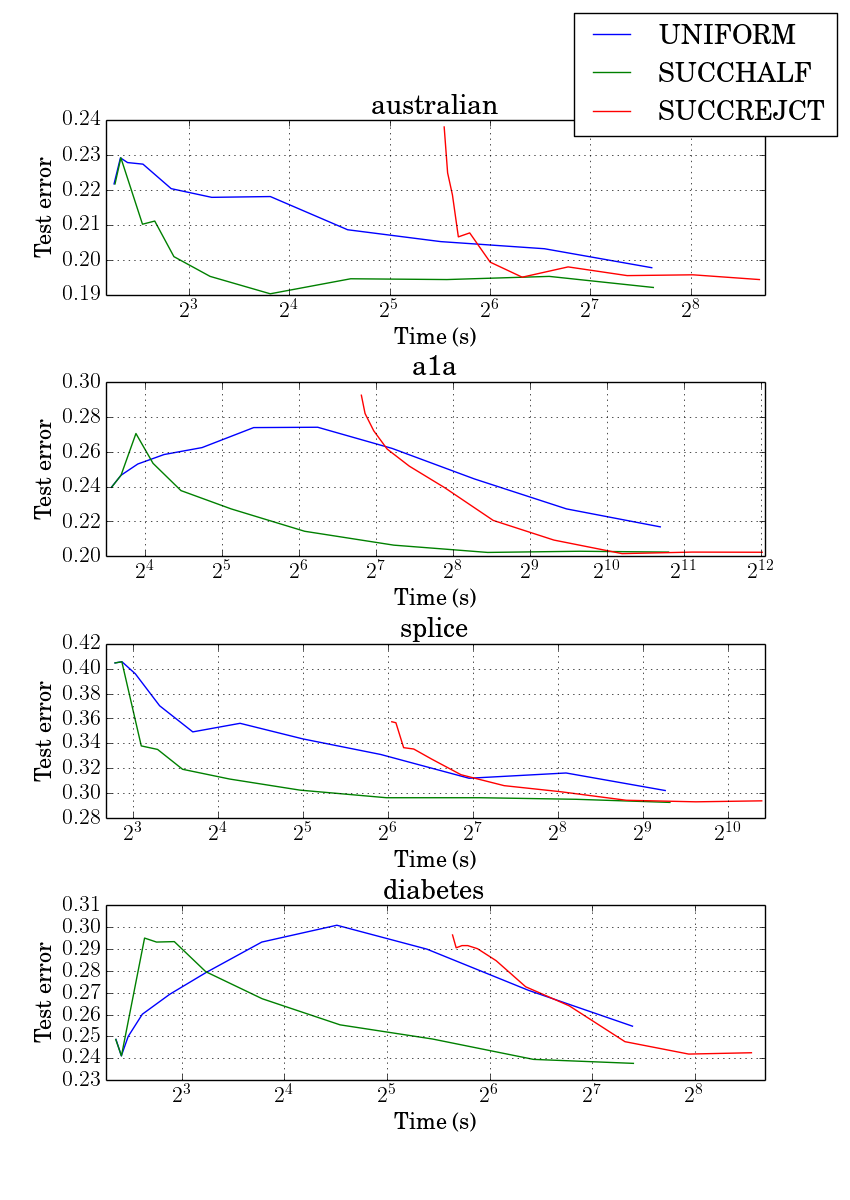}
\vspace{-.25in}
\caption{Kernel SVM. Successive Halving and Successive Rejects are
separated by an order of magnitude in wall-clock time.}
 \label{fig:krnsvm}
  \vspace{-.2in}
\end{figure}

We now consider learning a kernel SVM using the RBF kernel
$\kappa_{\gamma}(x,z) = e^{-\gamma ||x- z||_2^2}$. The SVM is trained using
Pegasos \cite{shalev2011pegasos} with
$\ell_2$ penalty hyperparameter $\lambda \in [10^{-6},10^0]$ and kernel width
$\gamma \in [10^{0},10^3]$ both chosen uniformly at random on a log scale per
trial. Each hyperparameter was allocated $10$ samples resulting in $10^2 = 100$
total arms. The experiment was repeated for 64 trials. Error on the $\VAL$ and
$\TEST$ was calculated using $0/1$ loss. Kernel evaluations were computed
online (i.e. not precomputed and stored). We observe in Figure~\ref{fig:krnsvm}
that Successive Halving obtains the same low error more than an order of
magnitude faster than both uniform and Successive Rejects with respect to
wall-clock time, despite Successive Halving and Success Rejects performing
comparably in terms of iterations (not plotted). 

 \begin{figure}
\centering
\includegraphics[width=.40\textwidth]{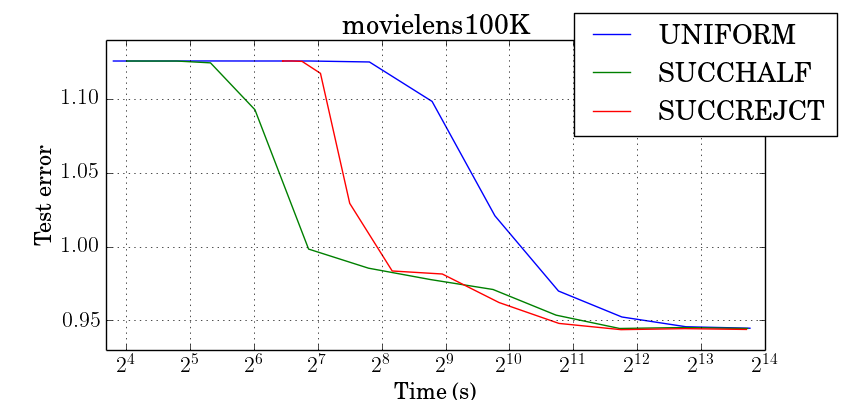} 
\caption{Matrix Completion (bi-convex formulation).}
 \label{fig:cf}
 \vspace{-.2in}
\end{figure}

\subsection{Collaborative filtering}
We next consider a matrix completion problem using the Movielens 100k dataset
trained using stochastic gradient descent on the bi-convex objective with step
sizes as described in \citet{recht2013parallel}. To account for the non-convex
objective, we initialize the user and item variables 
with entries drawn from a normal distribution with variance
$\sigma^2 / d$, hence each arm has hyperparameters $d$ (rank), $\lambda$
(Frobenium norm regularization), and $\sigma$ (initial conditions). $d \in
[2,50]$ and $\sigma \in [.01,3]$ were chosen uniformly at random from a linear
scale, and $\lambda \in [10^{-6},10^0]$ was chosen uniformly at random on a log
scale. Each hyperparameter is given 4 samples resulting in $4^3 = 64$ total
arms. The experiment was repeated for 32 trials. Error on the $\VAL$ and
$\TEST$ was calculated using mean-squared-error.  One observes in
Figure~\ref{fig:cf} that the uniform allocation takes two to eight times longer
to achieve a particular error rate than Successive Halving or Successive
Rejects.

\section{Future directions}

Our theoretical results are presented in terms of $\max_i \gamma_{i}(t)$. An
interesting future direction is to consider algorithms and analyses that take
into account the specific convergence rates $\gamma_i(t)$ of each arm,
analogous to considering arms with different variances in the stochastic case
\cite{kaufmann2014complexity}. Incorporating pairwise switching costs into the
framework could model the time of moving very large intermediate models in and
out of memory to perform iterations, along with the degree to which resources
are shared across various models (resulting in lower switching costs).
Finally, balancing solution quality and time by adaptively sampling
hyperparameters as is done in Bayesian methods is of considerable practical
interest.



\clearpage
\bibliography{autotuning_icml}
\bibliographystyle{icml2015}

\clearpage
\appendix
\onecolumn 

\section{Proof of Theorem~\ref{adaptive}}\label{adaptive-proof}

\begin{proof}
For notational ease, define $[ \cdot ] = \{ \{ \cdot \}_{t=1} \}_{i=1}^n$ so that $[ \ell_{i,t} ] = \{ \{ \ell_{i,t} \}_{t=1}^{\infty} \}_{i=1}^n$. Without loss of generality, we may assume that the $n$ infinitely long loss sequences $[ \ell_{i,t} ]$ with limits $\{ \nu_i \}_{i=1}^n$ were fixed prior to the start of the game so that the $\gamma_i(t)$ envelopes are also defined for all time and are fixed. Let $\Omega$ be the set that contains all possible sets of $n$ infinitely long sequences of real numbers with limits $\{ \nu_i \}_{i=1}^n$ and envelopes $[ \bar{\gamma}(t) ]$, that is,
\begin{align*}
\Omega = \left\{  [ \ell_{i,t}' ] :  [ \ | \ell_{i,t}' - \nu_i | \leq \bar{\gamma}(t) \ ] \wedge  \ \lim_{\tau \rightarrow \infty} \ell_{i,\tau}' = \nu_i \ \ \ \forall i \  \right\}
\end{align*}
where we recall that $\wedge$ is read as ``and'' and $\vee$ is read as ``or.''
Clearly, $[ \ell_{i,t} ]$ is a single element of $\Omega$. 

We present a proof by contradiction. We begin by considering the singleton set
containing $[\ell_{i,t} ]$ under the assumption that the Successive Halving
algorithm fails to identify the best arm, i.e., $S_{\lceil \log_2(n) \rceil}
\neq 1$.  We then consider a sequence of subsets of $\Omega$, with each one contained
in the next.
The proof is completed by showing that the final subset in our sequence (and
thus our original singleton set of interest) is empty when $B > z$, which contradicts
our assumption and proves the statement of our theorem.

To reduce clutter in the following arguments, it is understood that $S_k'$ for
all $k$ in the following sets is a function of $[ \ell_{i,t}' ]$ in the sense
that it is the state of $S_k$ in the algorithm when it is run with losses $[
\ell_{i,t}' ]$. We now present our argument in detail, starting with the singleton set of interest, and using the definition of $S_k$ in Figure~\ref{succHalfAlg}. 
\begin{align}
&\bigg\{   [ \ell_{i,t}' ] \in \Omega :  [ \ell_{i,t}'=\ell_{i,t} ]   \wedge  S_{\lceil \log_2(n) \rceil}'  \neq 1 \bigg\} \nonumber \\
&= \bigg\{   [ \ell_{i,t}' ] \in \Omega  :  [ \ell_{i,t}'=\ell_{i,t} ]  \wedge \bigvee_{k=1}^{\lceil \log_2(n) \rceil } \{ 1 \notin S_{k}'  , \ 1 \in S_{k-1}' \} \bigg\} \nonumber \\
&= \bigg\{   [ \ell_{i,t}' ] \in \Omega  :  [ \ell_{i,t}'=\ell_{i,t} ] \wedge \bigvee_{k=0}^{\lceil \log_2(n) \rceil -1 } \bigg\{  \sum_{i \in S_k'} \1\{  \ell_{i,R_k}' < \ell_{1,R_k}' \} > \lfloor |S_k'| /2 \rfloor \bigg\} \bigg\} \nonumber \\
&= \bigg\{   [ \ell_{i,t}' ] \in \Omega  :  [ \ell_{i,t}'=\ell_{i,t} ]  \wedge\bigvee_{k=0}^{\lceil \log_2(n) \rceil -1 } \bigg\{  \sum_{i \in S_k'} \1\{  \nu_i - \nu_1  < \ell_{1,R_k}'-\nu_1 - \ell_{i,R_k}' + \nu_i  \} > \lfloor |S_k'| /2 \rfloor \bigg\} \bigg\} \nonumber \\
&\subseteq \bigg\{   [ \ell_{i,t}' ] \in \Omega  :   \bigvee_{k=0}^{\lceil \log_2(n) \rceil -1} \bigg\{  \sum_{i \in S_k'} \1\{  \nu_i - \nu_1  < | \ell_{1,R_k}'-\nu_1| + |\ell_{i,R_k}' - \nu_i|  \} > \lfloor |S_k'| /2 \rfloor \bigg\} \bigg\} \nonumber \\
&\subseteq \bigg\{   [ \ell_{i,t}' ] \in \Omega  :  \bigvee_{k=0}^{\lceil \log_2(n) \rceil -1} \bigg\{  \sum_{i \in S_k'} \1\{ 2 \bar{\gamma}(R_k) >  \nu_i - \nu_1    \} > \lfloor |S_k'| /2 \rfloor \bigg\} \bigg\} \label{preOmegaBar} \,,
\end{align}
where the last set relaxes the original equality condition to just considering
the maximum envelope $\bar{\gamma}$ that is encoded in $\Omega$. The summation in Eq.~\ref{preOmegaBar} only involves the $\nu_i$, and this summand is maximized if each $S_k'$ contains the first $|S_k'|$ arms.  Hence we have,
\begin{align}
\eqref{preOmegaBar}&\subseteq \bigg\{   [ \ell_{i,t}' ] \in \Omega  :  \bigvee_{k=0}^{\lceil \log_2(n) \rceil -1 } \bigg\{  \sum_{i =1}^{|S_k'|} \1\{ 2 \bar{\gamma}(R_k) >  \nu_i  - \nu_{1}  \} > \lfloor |S_k'|/2 \rfloor \bigg\} \bigg\} \nonumber \\ 
&= \bigg\{   [ \ell_{i,t}' ] \in {\Omega}  :   \bigvee_{k=0}^{\lceil \log_2(n) \rceil -1 } \left\{  2\bar{\gamma}(R_k) >  \nu_{ \lfloor |S_k'|/2 \rfloor +1} - \nu_{1}   \right\} \bigg\} \nonumber \\
&\subseteq \bigg\{   [ \ell_{i,t}' ] \in {\Omega}  :  \bigvee_{k=0}^{\lceil \log_2(n) \rceil -1} \left\{  R_k < \bar{\gamma}^{-1} \left(\tfrac{\nu_{ \lfloor |S_k'|/2 \rfloor + 1 }- \nu_{1}}{2} \right)  \right\} \bigg\} \,,\label{postOmegaBar}
\end{align}
where we use the definition of $\gamma^{-1}$ in Eq.~\ref{postOmegaBar}. 
Next, we recall that $R_k = \sum_{j=0}^k \lfloor \frac{B}{ |S_k| \lceil \log_2(n) \rceil } \rfloor \geq  \frac{B / 2 }{ ( \lfloor |S_k| /2 \rfloor + 1 )\lceil \log_2(n) \rceil } -1$ since $|S_k| \leq 2 ( \lfloor |S_k|/2 \rfloor + 1 )$.  We note that we are underestimating by almost a factor of $2$ to account for integer effects in favor of a simpler form. By plugging in this value for $R_k$ and rearranging
we have that 
\begin{align*}
\eqref{postOmegaBar} &\subseteq \bigg\{   [ \ell_{i,t}' ] \in {\Omega}  :  \bigvee_{k=0}^{\lceil \log_2(n) \rceil -1} \left\{   \tfrac{B / 2}{ \lceil \log_2(n) \rceil  } < ( {  \lfloor |S_k'|/2 \rfloor } + 1 ) ( 1 + \bar{\gamma}^{-1} \left(\tfrac{ \nu_{ \lfloor |S_k'|/2 \rfloor + 1  } - \nu_{1}  }{2} \right) )  \right\} \bigg\} \\
&= \bigg\{   [ \ell_{i,t}' ] \in {\Omega}  :   \tfrac{B/2}{ \lceil \log_2(n) \rceil  } < \max_{k=0,\dots,\lceil \log_2(n) \rceil-1 } ( {  \lfloor |S_k'|/2 \rfloor } + 1 )(1 + \bar{\gamma}^{-1} \left(\tfrac{ \nu_{ \lfloor |S_k'|/2 \rfloor + 1 } - \nu_{1}  }{2} \right) )   \bigg\} \\
&\subseteq \bigg\{   [ \ell_{i,t}' ] \in {\Omega}  :    B < 2 \lceil \log_2(n) \rceil   \, \max_{i=2,\dots,n}  i \, ( \bar{\gamma}^{-1} \left(\tfrac{ \nu_{i} - \nu_1 }{2} \right) + 1)   \bigg\}  = \emptyset
\end{align*}
where the last equality holds if  $B > z$. 

The second, looser, but perhaps more interpretable form of $z$ is thanks to \cite{audibert2010best} who showed that
\begin{align*}
 \max_{i=2,\dots,n}  i \, \bar{\gamma}^{-1} \left(\tfrac{ \nu_{i} - \nu_1 }{2} \right)
&\leq \sum_{i=2,\dots,n}  \bar{\gamma}^{-1} \left(\tfrac{ \nu_{i} - \nu_1 }{2} \right) \leq \log_2(2n)  \, \max_{i=2,\dots,n}  i \, \bar{\gamma}^{-1} \left(\tfrac{ \nu_{i} - \nu_1 }{2} \right) 
\end{align*}
where both inequalities are achievable with particular settings of the $\nu_i$ variables. 
\end{proof}

\section{Proof of Theorem~\ref{nonadaptive}}
\begin{proof}
Recall the notation from the proof of Theorem~\ref{adaptive} and let $\widehat{i}([\ell_{i,t}'])$ be the output of the uniform allocation strategy with input losses $[\ell_{i,t}']$.
\begin{align*}
\bigg\{   [ \ell_{i,t}' ] \in \Omega :  [ \ell_{i,t}'=\ell_{i,t} ]   \wedge  \widehat{i}([\ell_{i,t}']) \neq 1 \bigg\} &=  \bigg\{   [ \ell_{i,t}' ] \in \Omega :  [ \ell_{i,t}'=\ell_{i,t} ]   \wedge \ell_{1,B/n}' \geq \min_{i=2,\dots,n}  \ell_{i,B/n}' \bigg\} \\
&\subseteq \bigg\{   [ \ell_{i,t}' ] \in \Omega  :   2\bar{\gamma}(B/n) \geq \min_{i=2,\dots,n}  \nu_i - \nu_1 \bigg\}\\ 
&=   \bigg\{   [ \ell_{i,t}' ] \in \Omega  : 2\bar{\gamma}(B/n) \geq  \nu_2 - \nu_1 \bigg\}\\ 
&\subseteq   \bigg\{   [ \ell_{i,t}' ] \in \Omega :    B \leq n  \bar{\gamma}^{-1}\left( \tfrac{\nu_2 - \nu_1}{2} \right) \bigg\} = \emptyset
\end{align*}
where the last equality follows from the fact that $B > z$ which implies $ \widehat{i}([\ell_{i,t}])= 1$.
\end{proof}

\section{Proof of Theorem~\ref{nonadaptive_necessity}}
\begin{proof}
Let $\beta(t)$ be an arbitrary, monotonically decreasing function of $t$ 
with $\lim_{t \rightarrow \infty} \beta(t) = 0$.  Define $\ell_{1,t} = \nu_1 +
\beta(t)$ and $\ell_{i,t} = \nu_i - \beta(t)$ for all $i$. Note that for all
$i$, $\gamma_i(t) = \bar{\gamma}(t) =  \beta(t)$ so that
\begin{align*}
 \widehat{i} = 1   &\iff    \ell_{1,B/n} < \min_{i=2,\dots,n}  \ell_{i,B/n}  \\
&\iff \nu_1 + \bar{\gamma}(B/n) < \min_{i=2,\dots,n}  \nu_i - \bar{\gamma}(B/n)    \\
&\iff \nu_1 + \bar{\gamma}(B/n) < \nu_2 - \bar{\gamma}(B/n)   \\
&\iff \bar{\gamma}(B/n) < \frac{\nu_2 - \nu_1}{2}   \\
&\iff  B \geq n \bar{\gamma}^{-1}\left( \tfrac{\nu_2 - \nu_1}{2} \right)  .
\end{align*}
\end{proof}

\section{Proof of Theorem~\ref{thm:prettygood}}
We can guarantee for the Successive Halving algorithm of
Figure~\ref{succHalfAlg} that the output arm $\widehat{i}$ satisfies
\begin{align*}   
\nu_{\widehat{i}} - \nu_1 &= \min_{i \in S_{\lceil \log_2(n) \rceil}}  \nu_i - \nu_1\\
&= \sum_{k=0}^{\lceil \log_2(n) \rceil -1 }  \min_{i \in S_{k+1}} \nu_i  - \min_{i \in S_{k}} \nu_i\\
&\leq \sum_{k=0}^{\lceil \log_2(n) \rceil -1 }  \min_{i \in S_{k+1}} \ell_{i,R_k} - \min_{i \in S_{k}}  \ell_{i,R_k}  + 2 \bar{\gamma}(R_k)\\
&= \sum_{k=0}^{\lceil \log_2(n) \rceil -1 }2 \bar{\gamma}( R_k ) \leq \lceil\log_2(n)\rceil 2 \bar{\gamma}\left(  \lfloor \tfrac{ B }{n \lceil \log_2(n) \rceil } \rfloor \right)
\end{align*}
simply by inspecting how the algorithm eliminates arms and plugging in a trivial lower bound for $R_k$ for all $k$ in the last step.

\end{document}